\documentclass[10pt]{article}
\usepackage[margin=0.75in]{geometry}
\usepackage[authoryear,round]{natbib}
\usepackage{graphicx}
\usepackage{booktabs}
\usepackage{tabularx}
\usepackage{caption}
\usepackage{placeins}
\usepackage{xcolor}
\definecolor{linkblue}{HTML}{006A9E}
\usepackage[colorlinks=true,allcolors=linkblue,breaklinks=true]{hyperref}
\usepackage{url}
\hypersetup{pdfauthor={},pdftitle={Measuring the Cost of Conversational Memory in AI Agents}}
\newcommand{\retentionfigurewidth}{0.88\linewidth}

\title{How Much Conversation History Does an AI Agent Need?}
\author{
Danny Kim\\
Pembroke Pines Charter High School\\
Boca Raton, FL\\
\texttt{dkim0603@pinescharter.net}
\and
Fernando Koch, Ph.D.\\
Department of Electrical Eng and Computer Sciences\\
Florida Atlantic University\\
Boca Raton, FL\\
\texttt{kochf@fau.edu}
}
\date{}

\begin{document}
\maketitle
\begin{abstract}
How does using less conversational history change reader-token use and benchmark scores? We examine this question on 1,100 BEAM questions. We held the retrieval ranking and reader constant across five fixed passage budgets, and mean assigned scores rose as the passage budget grew. The score gain per additional token became smaller at the upper tested budgets. Answerable questions generally received higher scores with more context. Pooled abstention scores sometimes rose as the passage budget shrank. We also tested message retention in a controlled task with explicit conversational-state targets and compared retained user messages with token-matched recency. Qwen matched the full state more often for direct requirements when user messages were retained. Its full-state matching declined when answers depended on references to earlier offers. Mistral produced near-floor results across the retention conditions. A scoring audit leaves the correctness of BEAM answers unresolved. These experiments characterize the token and score consequences of the tested memory budgets. Reliable budget selection depends on retaining the evidence each question requires and validating the measure used to assess answer quality.
\end{abstract}


\begin{center}
\begin{minipage}{\linewidth}
\includegraphics[width=\linewidth]{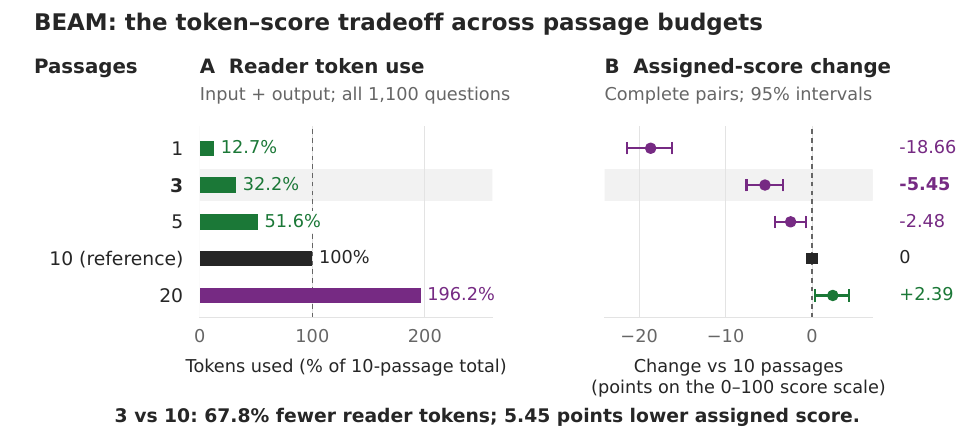}
\captionof{figure}{The BEAM cost--score tradeoff with fixed retrieval and a Qwen2.5-14B reader. \textbf{A:} Input-plus-output tokens for all 1,100 questions per budget, relative to ten passages (100\% = 7.98 million tokens); retrieval, preparation and judging are excluded. \textbf{B:} Paired assigned-score changes from ten passages, with 95\% percentile bootstrap intervals assuming 53 exchangeable operational source groups in this fixed cohort; these do not measure model-seed variation. Complete pairs number 1,087, 1,084, 1,087 and 1,085 for 1, 3, 5 and 20 passages; the reference is zero by definition. The shaded row highlights the primary 3-versus-10 comparison; other budgets are exploratory. Rubric scores are not percentages of correct answers, and we have not independently validated answer correctness.}
\label{fig:overview}
\end{minipage}
\end{center}
\clearpage

\section{Introduction}
\label{sec:intro}

Existing benchmarks, including LoCoMo, LongMemEval, and BEAM, evaluate whether agents can retrieve and use evidence from earlier conversations~\citep{maharana2024locomo,wu2025longmemeval,tavakoli2026beam}. They establish long-conversation memory as a measurable capability and show that relevant historical evidence affects performance. Research on retrieval budgets and adaptive context selection examines ways to reduce the text sent to a reader~\citep{tavakoli2026beam,taguchi2025adaptivek}. Research on automated evaluation reports imperfect alignment with human judgment and vulnerabilities in language-model judging~\citep{thakur2025judging,dietz2025principles}.


However, there is an open question: \emph{How much history can a reader lose before token savings affect the evidence available for an answer?} Agents use stored conversation history to answer questions about earlier interactions. A retrieval system selects passages from that history and supplies them to a language-model \emph{reader}. The system needs enough evidence to support a reliable answer and enough restraint to keep reader-token use manageable. We explore the following questions:
\begin{itemize}
    \setlength{\itemsep}{1pt}
    \item How does retrieving fewer passages change reader-token use and the score assigned to an answer?
    \item Does this tradeoff differ between questions that can be answered and questions that require abstention?
    \item What can an assigned benchmark score establish about correctness with respect to the source conversation?
    \item Does a message-retention policy keep the evidence needed to answer a question?
\end{itemize}


We study how a smaller memory budget changes reader cost and the evidence left for an answer. The fixed-budget experiment varies the passages passed to the reader for the same BEAM questions. It holds the retrieval ranking and reader configuration constant so that the passage budget drives the observed differences in reader-token use and assigned scores. We also review the scoring process to see what those scores can tell us about answer quality. The retention experiment turns to dependencies between messages. It asks whether a selective retention policy preserves both direct requirements and references to earlier offers. Together, these experiments let us compare token savings with the evidence that remains and the score the benchmark assigns to an answer.


This paper contributes to the state of the art by introducing:
\begin{itemize}
    \setlength{\itemsep}{1pt}
    \item An experimentation framework for measuring how passage budgets affect reader-token cost and assigned scores within a fixed retrieval-and-reading pipeline.
    \item An analysis of how smaller budgets affect answerable and abstention-labeled questions, and what their benchmark scores reveal about answer quality.
    \item A controlled test of message retention that compares direct requirements with references to earlier messages.
\end{itemize}

\section{Related work}
\label{sec:background}


We analyse existing work on how agents retrieve conversational history, reduce the context sent to a reader, use available evidence, and evaluate the resulting answer.


Memory benchmarks provide a common setting for this work: agents must recover evidence from earlier conversations and use it to answer a later question. LoCoMo evaluates this ability through question answering, summarization, and multimodal dialogue on human-verified long conversations~\citep{maharana2024locomo}. LongMemEval carries the task across extraction, temporal and multi-session reasoning, updates, and abstention, as well as indexing, retrieval, and reading~\citep{wu2025longmemeval}. BEAM examines the same problem over longer histories with human-validated questions, annotator-created rubrics, and retrieved-document budgets~\citep{tavakoli2026beam}. These benchmarks make memory performance observable, and their human validation supports the quality of questions and rubrics. Validating a generated answer against its source history requires a separate check.

The benchmark task depends on the history a system makes available to its reader. Memory architectures determine that history by deciding what to store, retrieve, and retain. Generative Agents combines experience storage, retrieval, and reflection to support agent behavior~\citep{park2023generative}, and MemGPT manages memory tiers for extended conversations and document analysis~\citep{packer2023memgpt}. At the passage-selection level, Adaptive-$k$ selects a passage count from retrieval similarities~\citep{taguchi2025adaptivek}. At the token level, LLMLingua reduces prompt length through budget control and token-level compression~\citep{jiang2023llmlingua}.

Making evidence available does not guarantee that a reader can use it. For instance, \citet{liu2024lost} show that evidence position affects reader performance. Sufficient Context distinguishes evidence that is missing from evidence that a reader fails to use, and it examines guided abstention~\citep{joren2025sufficient}. This distinction becomes more demanding in conversation, where a later message may depend on an earlier offer, commitment, or condition. Related work develops this issue from other directions. \citet{setiawan2026grab} tests abstention with missing and misleading passages. \citet{shen2026eviction} restores gold evidence to separate eviction losses from retrieval failures. \citet{spencer2026groundtruth} generates time-bounded facts before rendering conversations and compares memory architectures over time. These studies frame retention as a problem of preserving links between messages as well as the messages themselves.


Therefore, the gap lies in understanding how a smaller passage budget changes reader-token use and the evidence available for an answer within one fixed pipeline. This study addresses that gap with controlled passage-budget and message-retention experiments, paired score comparisons, and a scoring audit.

\section{Experiments}
\label{sec:method}


We introduce an experimentation framework for examining how conversational-memory choices change reader cost, available evidence, and assigned scores.


\subsection{Working with fixed passage budgets}
\label{sec:budget_experiment}

In this experiment, we vary the number of passages supplied to the same reader while holding the questions, retrieval ranking, and scoring procedure fixed. This paired design measures the cost--score tradeoff within one pipeline.

To measure this tradeoff, we use BEAM, which contains synthetic histories, human-validated probing questions, and annotator-created rubrics~\citep{tavakoli2026beam}. We use all questions from fixed chat IDs 1--20 in its 100K directory and 1--35 in its 500K directory: 400 and 700 questions, respectively. The dataset revision is \texttt{b2da22ea}; the full revision and model pins are in the anonymous supplement, \texttt{protocol/beam.json}. The smaller tier is called 128K upstream. These labels describe stored histories, not prompt lengths. Ten categories contribute equally within each tier; 990 questions are labeled answerable and 110 require abstention. The cohort, budgets, retrieval, reader, scorer, main comparisons, source grouping, and review sample were fixed before answer generation.

BM25 ($k_1=1.2$, $b=0.75$) ranks 1,200-character chunks with 160-character overlap and one neighboring chunk on either side. The ranking preserves dates, speakers, and message boundaries, and removes repeated source characters. The reader receives the first 1, 3, 5, 10, or 20 passages from that ranking. Two rankings contain only 12 and 18 passages. The pinned Qwen2.5-14B-Instruct reader~\citep{qwen2025report} uses greedy decoding, seed zero, a 32,768-token window, and a 1,024-token output limit. It is instructed to answer ``No information available.'' when the supplied evidence is insufficient.

Reader cost includes input and output tokens, with retrieval, preparation, and judging excluded. BEAM's rubric assigns each criterion 0, 0.5, or 1. We average criteria within questions and then average equally across questions on a 0--100 scale. The score is not the fraction of wholly correct answers. The original \texttt{gpt-4.1-mini-2025-04-14} judge uses temperature zero and a 1,024-token limit. Its prompt receives the criterion and candidate answer, but not the question or full history. Invalid JSON remains missing without repair or retries. Gold answers and rubrics never enter reader prompts, and BEAM's event-ordering metric remains outside the study scope.

Our primary analysis estimates the mean paired score change from ten to three passages. We use source-group bootstrap intervals to account for shared conversational material across chats. Scores use complete pairs, token totals use all questions, and the remaining budget comparisons are exploratory.


\subsection{Comparing answerable and abstention-labeled questions}
\label{sec:answerability_analysis}

The fixed-budget experiment also asks whether the cost--score tradeoff changes with answerability. For that, an answerable question calls for evidence that supports a substantive answer. An abstention-labeled question calls for withholding an answer when the available evidence is insufficient. We compare the top-3/top-10 score effect between these groups within each tier. The interaction is the difference between the two group effects; positive values indicate that budget reduction favors abstention more than answerability. The planned interactions use 97.5\% source-group intervals, a Bonferroni adjustment that targets 95\% family coverage. Subgroup intervals are 95\%. The labels provide evaluation information and do not enter a budget selector. We also report exploratory paired means for all ten categories across the four budget contrasts.

\subsection{Checking the reliability of assigned scores}
\label{sec:scoring_audit}

The original judge receives criteria and candidate answers without the full history, so agreement with its scores alone cannot establish source-grounded correctness. We therefore review sampled references and answers, test a source-aware replacement judge, and assess the sensitivity of the primary score difference.

The preselected review sample contains 28 questions per tier: ten abstention questions and two from each answerable category. A Codex AI reviewer inspects source evidence and references before the sampled answers, then records ratings before budget labels and individual scores are joined. A source-aware replacement judge, \texttt{gpt-4.1-mini-2025-04-14}, must pass every declared source and answer check on each of 31 synthetic cases before it can rescore BEAM. The cases include 21 reused regressions and ten fresh probes; each requires one source call and one answer call. Post-hoc sensitivity checks remove each source group in turn and bound arbitrary changes to both scores on selected pairs. Missing pairs remain unrestricted. These checks measure the stability of the paired score difference.


\subsection{Tracing evidence through message retention}
\label{sec:retention_development}

The retention experiment tests whether selective message retention preserves the information needed to answer. It compares paired conversations in which a user either states a value directly or accepts a named earlier offer. The acceptance depends on retaining that offer, whereas the direct statement supplies the value itself. This design separates information loss from the reader's use of the visible text.

Each structural family fixes an ordered schedule of updates, offers, and value-equality relationships. Sixteen reserved families, selected in ID order, cross travel and reservation vocabularies, two timestamp wrappers, and direct or reference-dependent variants, producing 128 stimuli per reader. Timestamp wrappers share event sentences. Each trajectory has three initial assignments, 6--18 update or offer events, one item offer and acceptance, and 3--7 trailing offers. The direct counterpart replaces only the acceptance with an explicit assignment.

\begin{figure}[!htb]
\centering
\includegraphics[width=\linewidth]{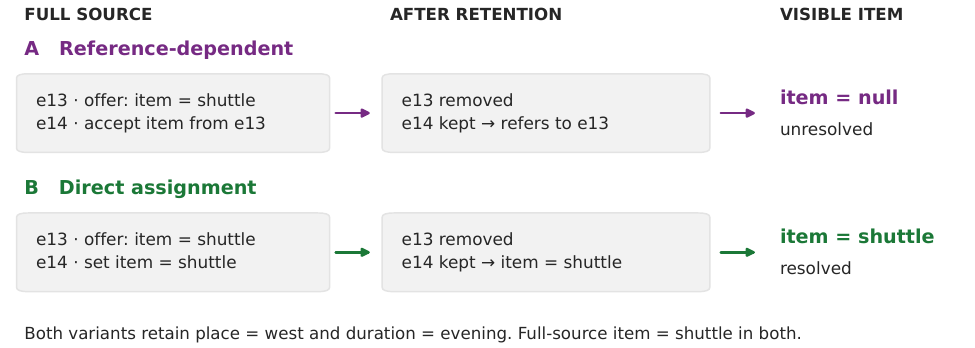}
\caption{A dependency removed by retention. The rows isolate the decisive messages in two versions of a frozen task; columns show the full source, retained text and visible item. \textbf{A:} e14 names offer e13, so removing e13 leaves the item unresolved. A later offer of the same value cannot resolve that reference. \textbf{B:} e14 states the value directly, so it survives. Both variants retain place and duration. Oracle restoration repairs the source dependency but also changes padding and position. This schematic describes task grammar; model outcomes are measured separately.}
\label{fig:design}
\end{figure}

The reader reconstructs \texttt{item}, \texttt{place}, and \texttt{duration}. The latest user requirement persists. An assistant offer changes nothing unless a user accepts its named slot. Unresolvable slots must return \texttt{null}. Figure~\ref{fig:design} shows a frozen example. Forward replay, a separate backward checker, and a grammar recognizer validate the targets and renderings.

The policies compare full history, user-history retention, and role-blind recency. User-history retention keeps user messages and the last three messages. Role-blind recency keeps a whole-message suffix at the same prompt-token count. Full-history guidance stresses persistent user requirements and accepted references; its placebo is a task-irrelevant collection note. Padding matches these prompts to the longer one. Two additional arms remove the decoder constraint from full history and retention.

Oracle restoration adds the named offer to retention in both variants in its original chronological order. The direct variant already states its value. Its removal control uses the same enlarged token count. All matching uses labeled nonfactual padding. Oracle selection uses evaluation evidence and is not deployable. Except for the unconstrained arms, decoding enforces exactly \texttt{\{\"state\":\{\"item\":v,\"place\":v,\"duration\":v\}\}}, where each \texttt{v} is an unrestricted string or \texttt{null}; no gold value is enumerated.

Pinned Qwen2.5-14B-Instruct and Mistral-7B-Instruct-v0.3 use vLLM 0.11.2, temperature and seed zero, and a 128-token output limit. Prompts have at most 1,077 tokens without clipping. The design has 2,304 study requests and four interface canaries.

Operational accuracy requires all three slots to match the \emph{full-source} state in compliant, completed JSON. A null value consistent with incomplete visible text can still lose full-source accuracy. Source sufficiency and response support are separate endpoints; response support remains unresolved when the visible reference cannot be resolved. Variants remain clustered within family. The comparisons are descriptive development results without held-out confirmation, and missing-outcome bounds are not confidence intervals.

\section{Results}
\label{sec:results}
Memory budgets determine both the reader tokens used for an answer and the evidence that reaches the reader. The results address the four questions in the introduction by measuring the token--score tradeoff, examining how it varies across question types, auditing the meaning of an assigned score, and testing retention under known message dependencies.

\subsection{Reader-token use and assigned scores}
\label{sec:analysis_budget_score}

\emph{How does retrieving fewer passages change reader-token use and the score assigned to an answer?}

The fixed-budget experiment shows a measurable exchange between reader-token use and assigned score. Reducing ten passages to three saves \textbf{67.78\% of reader tokens} with a paired loss of \textbf{5.45 score points} (95\% interval: 3.35--7.59 points of loss). Twenty passages use 96.24\% more tokens for a 2.39-point gain (Figure~\ref{fig:overview}). These results place the tested budgets on a cost--score tradeoff within the fixed retrieval-and-reading pipeline.

The completed pairs support the direction of this tradeoff. All 5,500 answers completed without prompt clipping; three reached the output limit. Judging yielded 5,457 scores and 43 JSON failures. The 1,084 primary complete pairs average 43.43 at top-3 and 48.88 at top-10, with negative unrestricted missing-pair bounds ($[-6.82,-3.91]$ points). At top-3, 124 pairs score higher, 699 equally and 261 lower; 284 score zero in both conditions. The question therefore has a clear answer for the tested conditions: fewer passages substantially reduce reader-token use and lower the mean assigned score.

\subsection{Answerability and the budget tradeoff}
\label{sec:analysis_answerability}

\emph{Does this tradeoff differ between questions that can be answered and questions that require abstention?}

\begin{figure}[!htb]
\centering
\includegraphics[width=\linewidth]{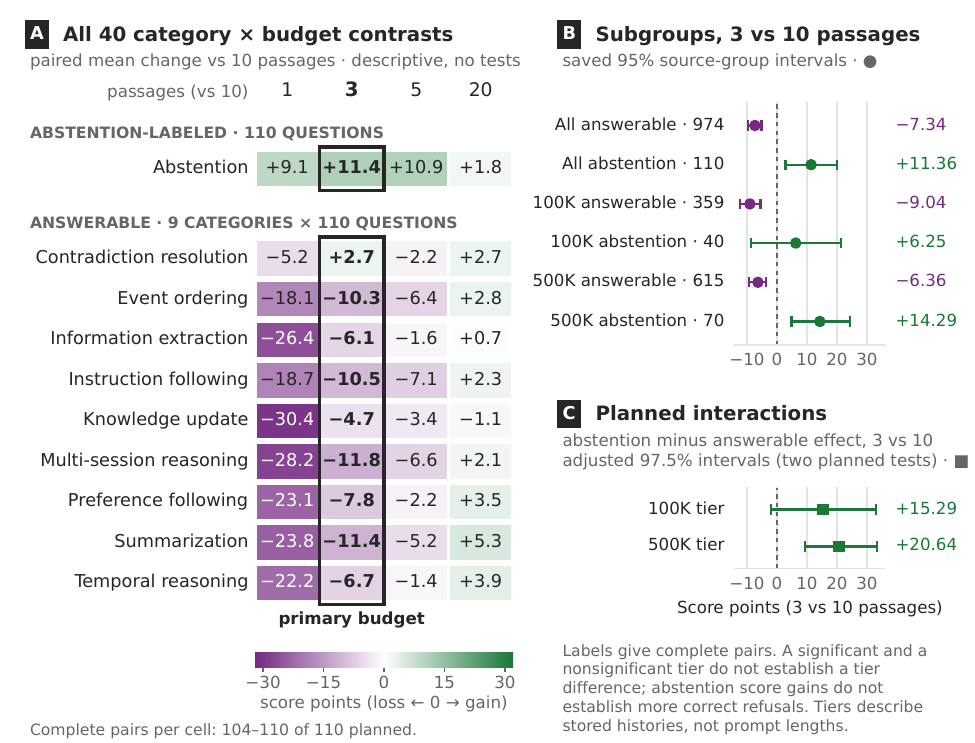}
\caption{Where passage budgets help and hurt assigned scores. \textbf{A:} All 40 category--budget contrasts as paired means on one zero-centered color scale; signed cell values give score-point changes from ten passages. Outlines mark top-3. Each cell uses 104--110 complete pairs from 110 planned questions. Categories are descriptive, not significance tests. \textbf{B:} Top-3/top-10 subgroup effects with saved 95\% source-group intervals; labels give paired counts. \textbf{C:} Planned interactions with adjusted 97.5\% intervals: 100K, $+15.29\;[-1.96,32.88]$; 500K, $+20.64\;[9.22,33.36]$ points. The tiers contain 20 and 33 source groups, respectively.}
\label{fig:answerability}
\end{figure}

The overall tradeoff masks different score patterns for answerable and abstention-labeled questions. Pooled abstention has positive point estimates at every smaller budget, while most answerable categories lose score (Figure~\ref{fig:answerability}A). The planned interaction excludes zero at 500K and includes zero at 100K (Figure~\ref{fig:answerability}C). The 500K result supports a difference in assigned-score effects between the two groups, whereas the 100K result remains uncertain. Neither result establishes more correct refusals.

\subsection{What benchmark scores establish}
\label{sec:analysis_scoring}

\emph{What can an assigned benchmark score establish about correctness with respect to the source conversation?}

\begin{figure}[!htb]
\centering
\includegraphics[width=\linewidth]{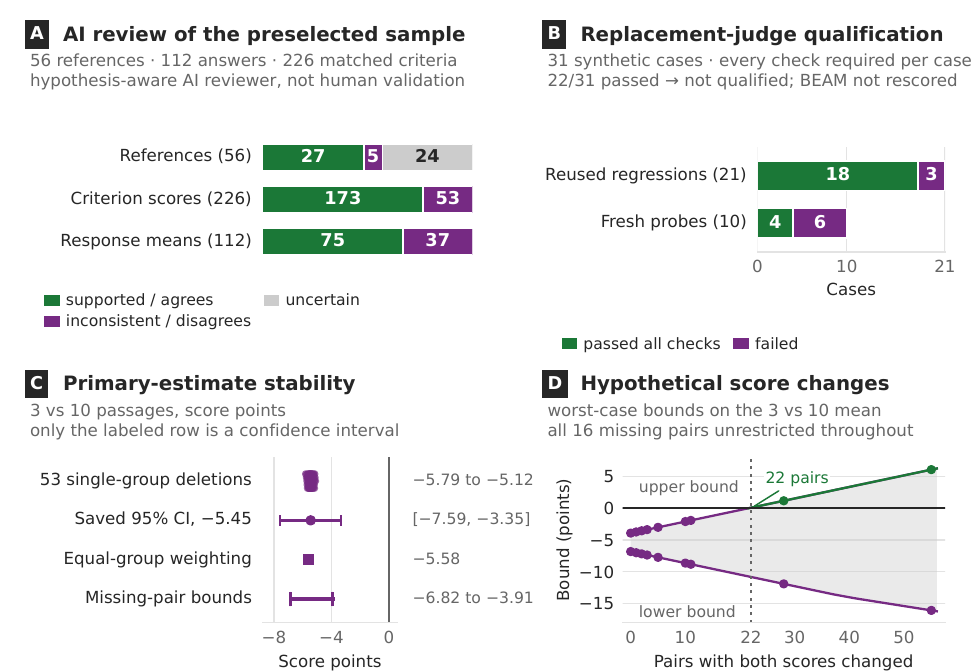}
\caption{Evaluation limits and score sensitivity. \textbf{A--B:} Selected AI-review outcomes and synthetic qualification cases passing all required checks. Passing cases comprise 18/21 reused regressions and 4/10 fresh probes; this is not an original-BEAM error-rate estimate. \textbf{C:} All 53 single-group deletion estimates, the original interval, equal-group weighting, and unrestricted missing-pair bounds. Only the row labeled 95\% is a confidence interval. \textbf{D:} Saved worst-case bounds when both scores on selected complete pairs may change arbitrarily; all 16 missing pairs are unrestricted throughout. The threshold of 22 pairs is a sensitivity result, not an observed error count or error rate.}
\label{fig:evaluation}
\end{figure}

The fixed-budget results establish differences in benchmark credit. The scoring audit leaves source correctness unresolved. The AI audit covers 56 references, 112 answers and 226 criterion comparisons (Figure~\ref{fig:evaluation}A--B). All 20 abstention references remain uncertain, and disagreement on 37 responses cannot establish evaluator correctness. The review also falls short of the planned independent, outcome-blind human review: the AI reviewer knew the hypothesis and aggregate results, searched selected history portions, and lacks an exact recorded model revision. The source-aware replacement judge did not pass every declared synthetic source and answer check, so original scores remain unchanged.

The audit results align with broader concerns about automated judging: \citet{thakur2025judging} document imperfect alignment with human judgment and vulnerabilities in language-model judges, and \citet{dietz2025principles} propose task-specific tests and safeguards for their use in information retrieval. Removing any single source group preserves the score loss, but arbitrary changes to 22 complete pairs can erase its sign under worst-case missing outcomes (Figure~\ref{fig:evaluation}C--D). The observed budget effects therefore describe assigned scores, with answer correctness remaining an open evaluation question.

Two cases show why the audit cannot turn an assigned score into a source-correctness measure. \texttt{beam-500K-22-q18} asks how many months separate a March 2024 portfolio value from a December 2026 goal. The AI audit flagged a reference/rubric mismatch: the prose says approximately 30 months; the rubric requires 33. The original answers and scores are:
\begin{center}
\small
\begin{tabularx}{\linewidth}{@{}lXr@{}}
\toprule
Budget & Saved answer, summarized & Assigned score \\
\midrule
3 passages & Identifies March 2024 to December 2026 as 33 months. & 100 \\
10 passages & Cannot determine the starting month; discusses the last 6 months. & 0 \\
\bottomrule
\end{tabularx}
\end{center}
This mismatch does not show original-judge misgrading. In \texttt{beam-100K-01-q00} (top-10), the rubric specifies no information about a UI/UX question, yet the judge scores ``No information available.'' zero because it expects positive details. This rubric-direction error does not estimate source correctness or error prevalence.

\subsection{Evidence preserved by message retention}
\label{sec:analysis_retention}

\emph{Does a message-retention policy keep the evidence needed to answer a question?}

\begin{figure}[!htb]
\centering
\includegraphics[width=\retentionfigurewidth]{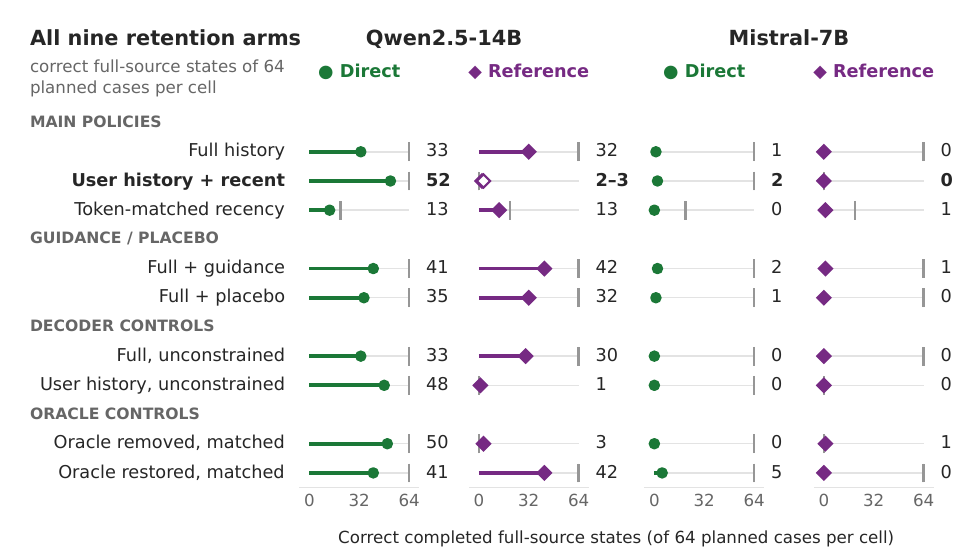}
\caption{All controlled retention outcomes on a common scale. Marks and counts show correct completed full-source states among 64 planned cases in 16 families; correct guesses need not be supported by retained text. Light ticks show source sufficiency, a property of retained text. The open diamond preserves one missing response; its range is not a confidence interval. Row groups separate the main policies, guidance/placebo, decoder and oracle controls. At matched tokens, oracle restoration changes Qwen correctness from 3/64 to 42/64 on reference cases and from 50/64 to 41/64 on direct cases. Position changes and padding prevent an isolated mechanism claim.}
\label{fig:retention}
\end{figure}

The controlled retention experiment tests this question with explicit state targets for direct requirements and references to earlier offers (Figure~\ref{fig:retention}, all arms and both readers). All 2,304 requests were attempted: 2,303 responses were compliant and untruncated; one Qwen reference-dependent retention request failed in transport without repetition. User-history retention preserves the complete state in 64/64 direct cases and 0/64 reference cases by design for both readers. Matched recency preserves it in 20/64 cases of each kind. The two policies therefore make different evidence available to the reader.

\begin{figure}[!htb]
\centering
\includegraphics[width=\retentionfigurewidth]{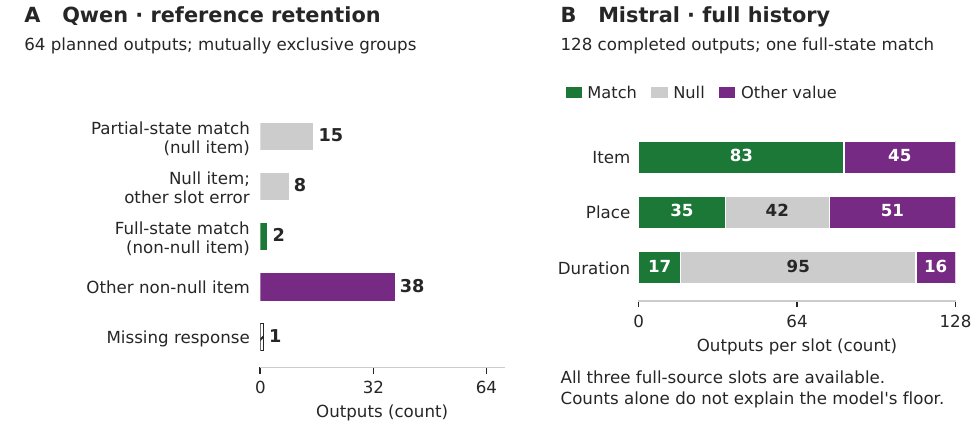}
\caption{Post-hoc diagnostics from saved responses. \textbf{A:} Qwen's 64 reference-retention requests partition into five groups. The 15 partial-state matches return a null item and correct visible place/duration; eight other null-item outputs miss at least one of those slots. Two full-state matches cannot establish support for the unresolved item. One response is missing. \textbf{B:} Mistral's full-history outputs, partitioned by slot into target match, null and other value. Only one output matches all three slots. These counts explain what the original all-slot metric combines; they do not replace it or establish a causal explanation.}
\label{fig:retention_diagnostics}
\end{figure}

Qwen matches the full state more often under retention than matched recency in 13/16 direct-case families. Reference effects are mixed and negative overall. Pooled retention beats matched recency and falls below full history. Figure~\ref{fig:retention_diagnostics} distinguishes partial-state from full-state matches and shows Mistral's slot-level failures. Support remains unresolved for incomplete references. Mistral's floor remains unexplained after removing decoder constraints and checking model identities and compiled requests, limiting inference about its response to retention.

The retention policy also reduces compiled prompt tokens by 30.08\% for Qwen and 31.53\% for Mistral across planned cases, including 422 planned tokens for the failed Qwen request (actual usage unknown). These input-only totals differ from BEAM's input-plus-output usage. The result supports user-history retention for Qwen's direct requirements, but it does not support the same policy for references to earlier offers. Mistral does not corroborate the Qwen result. The memory policy therefore needs to preserve the evidence required by the dependency at hand, and the score used to assess the resulting answer needs separate validation.

\FloatBarrier

\section{Conclusion}
\label{sec:conclusions}

This work aimed to answer the question: \emph{How much history can a reader lose before token savings affect the evidence available for an answer?} We concluded that a smaller memory budget creates a measurable cost--score tradeoff and that the evidence a question requires sets the limit of a safe reduction. In the BEAM pipeline tested here, moving from ten passages to three cuts reader-token use by 67.78\% and lowers the mean assigned score by 5.45 points. Larger budgets raise assigned scores, and the upper tested budgets yield smaller observed gains per additional token. The tested evidence does not select a single best budget.

It is interesting to observe that answerable and abstention-labeled questions follow different assigned-score patterns as passage budgets shrink. Most answerable categories lose score, whereas pooled abstention scores have positive point estimates at every smaller budget. The planned interaction supports a difference between these groups at 500K and remains uncertain at 100K. The scoring audit limits what these patterns can establish because it leaves answer correctness with respect to the source conversation unresolved. The observed score tradeoff therefore cannot identify the budget that produces the most correct answers.

It is also interesting to observe that message dependencies shape the evidence a retention policy must keep. Retaining user messages helps Qwen reconstruct direct requirements more often than token-matched recency. The pattern reverses when the answer depends on an earlier offer, and Mistral provides no corroborating benefit. This reversal ties policy choice to the dependency in a question: the policy must preserve the evidence that dependency requires. A fixed amount of recent text cannot provide that guarantee.

In conclusion, we characterized the cost--score tradeoff produced by the tested memory budgets. However, we must be aware that the study uses one BEAM reader and retriever, selected synthetic retention families, and assigned scores without independent source-aware answer validation. These conditions do not establish a universal budget rule.

Moving forward, research should begin with a source-grounded quality measure, including outcome-blind human review of answers and abstentions. With that measure in place, researchers can freeze a budget or retention policy and test it on fresh source groups, natural conversations, held-out dependency families, additional readers and retrievers, and simple and published baselines. The evaluation should also account for retrieval, selection, output, and monetary costs. These tests would show whether the observed savings support reliable per-question memory decisions.

\bibliographystyle{plainnat}
\bibliography{references}
\end{document}